\def\papertitle{Reinforcement learning with experience replay \\ and adaptation of action dispersion}

\documentclass{article}

\usepackage[preprint]{neurips_2022}

\usepackage[utf8]{inputenc} % allow utf-8 input
\usepackage[T1]{fontenc}    % use 8-bit T1 fonts
\usepackage{hyperref}       % hyperlinks
\usepackage{url}            % simple URL typesetting
\usepackage{booktabs}       % professional-quality tables
\usepackage{amsmath} 
\usepackage{amsfonts}       % blackboard math symbols
\usepackage{nicefrac}       % compact symbols for 1/2, etc.
\usepackage{microtype}      % microtypography
\usepackage{xcolor}         % colors
\usepackage{ulem}
\usepackage{graphicx} 
\usepackage{algorithm} 
\usepackage{algorithmic} 
\usepackage{multicol} 

\title{\papertitle} 

\author{%
  Paweł Wawrzyński%\thanks{Use footnote for providing further information about author (webpage, alternative address)---\emph{not} for acknowledging funding agencies.} 
  \\
  Institute of Computer Science\\
  Warsaw University of Technology\\
  Warsaw, Poland \\
  \texttt{pawel.wawrzynski@pw.edu.pl} \\
  \And
  Wojciech Masarczyk \\
  Institute of Computer Science\\
  Warsaw University of Technology\\
  Warsaw, Poland \\
  \texttt{wojciech.masarczyk.dokt@pw.edu.pl} \\
  \AND
  Mateusz Ostaszewski \\
  Institute of Computer Science\\
  Warsaw University of Technology\\
  Warsaw, Poland \\
  \texttt{mateusz.ostaszewski@pw.edu.pl} \\
}

\def\headneurips{}

\newcommand\afterbegindocument{
\maketitle
}

\usepackage{caption}

\def\state{s}
\def\stateSpace{\mathbb{S}}
\def\ctrl{a}
\def\ctrlSpace{\mathbb{A}}

\def\Cparam{\nu}
\def\Aparam{\theta}
 
\def\real{\mathbb R}

\def\pd{\partial}
 
\def\T{^{\scriptsize T}}

\def\comment#1{}

\def\eqref#1{(\ref{#1})}
\def\Beq#1\Eeq{\begin{equation}#1\end{equation}}
\def\Beqo#1\Eeqo{\begin{equation*}#1\end{equation*}}
\def\Beqs#1\Eeqs{\begin{align}#1\end{align}}
\def\Beqso#1\Eeqso{\begin{align*}#1\end{align*}}

\newcommand\wyciete[1]{}

\begin{document} 

\afterbegindocument

\begin{abstract}
Effective reinforcement learning requires a~proper balance of exploration and exploitation defined by the dispersion of action distribution. However, this balance depends on the task, the current stage of the learning process, and the current environment state. Existing methods that designate the action distribution dispersion require problem-dependent hyperparameters. In this paper, we propose to automatically designate the action distribution dispersion using the following principle: This distribution should have sufficient dispersion to enable the evaluation of future policies. To that end, the dispersion should be tuned to assure a sufficiently high probability (densities) of the actions in the replay buffer and the modes of the distributions that generated them, yet this dispersion should not be higher. This way, a~policy can be effectively evaluated based on the actions in the buffer, but exploratory randomness in actions decreases when this policy converges. The above principle is verified here on challenging benchmarks Ant, HalfCheetah, Hopper, and Walker2D, with good results. Our method makes the action standard deviations converge to values similar to those resulting from trial-and-error optimization. 
\end{abstract}

\section{Introduction}

%\pawel{Komentarz Pawła} \wojtek{Komentarz Wojtka} \mateusz{Komentarz Mateusza} 

In reinforcement learning (RL), \citep{2018sutton+1} a~problem is considered of an agent that makes subsequent actions in a~dynamic environment. The actions change the state of the environment, and depending on them, the agent receives numeric rewards. The agent learns to designate actions based on states to receive the highest rewards in the future.

To optimize its behavior, the agent needs to observe the consequences of different actions, i.e.; it needs to apply diverse actions in each state. Therefore, the agent uses a~stochastic policy to designate actions, i.e., each time the agent draws them from a~certain distribution conditioned on the state. The more dispersed this distribution is, the more experience the agent gathers, but the less likely it becomes that the agent gets to states that yield high rewards. This so-called exploration-exploitation trade-off is essential for efficient reinforcement learning. Despite a~lot of significant research, adaptive optimization of this trade-off is still an open problem. 

In this paper, we analyze the following approach to designating dispersion of action distribution, thereby quantifying the exploration-exploitation trade-off. We assume that the agent experience is stored in a~memory buffer of a~fixed size, and the policy changes at a~certain pace due to learning. The learning is driven mostly by actions representative for evaluated policies. Therefore, the current action distribution should be so dispersed for the currently taken actions to have sufficiently high density in future policies for effective evaluation and selection of these policies.

%Within our approach, agent experience is sufficient to evaluate the current policy. However, when the policy converges, the dispersion of the action distributions decreases. 
The contribution of the paper can be summarized in the following points: 
\begin{itemize} 
\item 
We analyze the evaluation of future policies as a~primary reason for exploration in RL. We propose a~way to quantify exploration to enable that evaluation. 
\item 
We introduce an RL algorithm that automatically designates a~dispersion of action distribution (the amount of exploration) in the trained policy. This dispersion is sufficient to evaluate the current policy but not larger. Hence, when the policy converges, the dispersion is suppressed.  
\item 
We present simulations that demonstrate the efficiency of the above algorithm on four challenging learning control problems: Ant, HalfCheetah, Hopper, and Walker2D.
\end{itemize} 

The rest of the paper is organized as follows. Section~\ref{sec:related-work} overviews literature related to the topic of this paper. The following section formulates the problem of designating the amount of randomness in an agent's policy that optimizes its behavior with RL. Section~\ref{sec:method} presents our approach to this problem. In Section~\ref{sec:experiments}, simulations are discussed in which our method is compared with RL algorithms PPO, SAC, and ACER. %Section~\ref{sec:discussion} discusses the experimental results.
The last section concludes the paper.

\section{Related work} 
\label{sec:related-work} 

\paragraph{Exploration in reinforcement learning.} 

Most existing reinforcement learning algorithms are designed to optimize policies with a fixed amount of randomness in actions. This amount is defined by a~quantity such as action standard deviation or the probability of an~exploratory action. Within the common approach to RL, this quantity is tuned manually. A simple approach to automatize this tuning is to train this quantity as one of the policy parameters. This approach was first introduced as a~part of the REINFORCE \citep{1991williams+1} algorithm and later applied in the Asynchronous Advantage Actor-Critic \citep{2016mnih+many} algorithm and the Proximal Policy Optimization \citep{2017schulman+4} algorithm. However, a~policy learned this way degrades to a deterministic one without hand-crafted regularization \citep{2016mnih+many}. This regularization is typically introduced as an entropy-based bonus term.

A different approach is to control exploration to increase state-space coverage. One way to achieve this goal is to reward the agent for visiting novel states. State novelty may be determined with a counting table \citep{2017tang+many} or estimated using environment dynamics prediction error \citep{2017pathak+3,2015stadie+3}. Another way is to maximize the expected difference between the current policy and past policies, thus increasing the coverage of the policies space \citep{2018hong+many}. 

The maximum entropy RL is a different approach to optimize exploration while keeping policy from degrading to a deterministic one \citep{1957jaynes,2008ziebart+1}. In this approach, the policy is optimized with regard to a~quality index that combines actual rewards and the entropy of the action distribution. The first off-policy maximum entropy RL algorithm was the Soft Actor-Critic (SAC) \citep{2018haarnoja+3}  algorithm. It is the most prominent RL method to tune the amount of exploration during its operation. Although the idea of achieving that by rewarding for the action distribution entropy was a~breakthrough, it is still heuristic and sometimes does not work. And even if it does, it still requires a~handcrafted coefficient, namely the weight of the entropy. The follow-up version of SAC \citep{2019haarnoja+many} tunes this coefficient to approach a target level with
gradient descent dynamically. This target level is set for the finite action space $\ctrlSpace$ to $-|\ctrlSpace|$, a value that works empirically on benchmark tasks but is not justified otherwise. Meta-SAC \citep{2020wang+1} tunes the target entropy value with a meta-gradient approach.

Existing techniques reduce the balancing of exploitation and exploration to balancing of rewards and entropy. That generally requires problem-dependent coefficients. Within our approach, we define an~independent criterion for the amount of exploration, thereby, in principle, avoiding problem-dependent settings. 

\paragraph{Efficient utilization of previous experience in RL} 

The fundamental Actor-Critic architecture of reinforcement learning was introduced in~\citep{1983barto+2}. Approximators were applied to this structure for the first time in~\citep{1998kimura+2}. To boost the efficiency of these algorithms, experience replay (ER) \citep{1992mahadevan+1} can be applied, i.e., storing the events in a~database, sampling, and using them for policy updates several times per each actual event. ER was combined with the Actor-Critic architecture for the first time in~\citep{2009wawrzynski}. 

Application of the experience replay to Actor-Critic creates the following problem. The learning algorithm needs to estimate the quality of a~given policy based on the consequences of actions registered when a~different policy was in use. Importance sampling estimators are designed to do that, but they can have arbitrarily large variances. In~,\citep{2009wawrzynski} that problem was addressed with truncating density ratios present in those estimators. In~\citep{2016wang+6} specific correction terms were introduced for that purpose. 

% redo - following paragraph is about on-policy algorithms
The significance of the relevance of samples was noted for the on-policy algorithms that use experience buffer. This class of algorithms shows another approach to the problem above. They prevent the algorithm from inducing a~policy that differs too much from the one used to collect samples. That idea was first applied in Conservative Policy Iteration~\citep{2002kakade+1}. It was further extended in Trust Region Policy Optimization~\citep{2015schulman+4}. This algorithm optimizes a~policy with the constraint that the Kullback-Leibler divergence between that policy and the tried one should not exceed a~given threshold. The K-L divergence becomes an~additive penalty in Proximal Policy Optimization algorithms, namely PPO-Penalty and PPO-Clip~\citep{2017schulman+4}. 

A~way to avoid the problem of estimating the quality of a~given policy based on the tried one is to approximate the action-value function instead of estimating the value function. Algorithms based on this approach are Deep Q-Network (DQN) \citep{2013mnih+6}, Deep Deterministic Policy Gradient (DDPG) \citep{2016lillicrap+7}, and Soft Actor-Critic (SAC) \citep{2018haarnoja+3}. 
%In the original version of DDPG the time-correlated noise was added to action. However, this algorithm was not adapted to this fact in any specific way. 
SAC uses noise as input for calculating policy, and it is considered one of the most efficient in this family of algorithms. 

%As the efficiency of policy optimization is crucial for RL, the problem of proper selection of experience for their replay draws a~lot of attention~\citep{2018wan+1,2020tan+3}. Also, optimization of the~$\lambda$ parameter for the $SARSA(\lambda)$ algorithm with ER is considered in \citep{2019deley+1}. %However, similar enhancements in the more efficient and difficult Actor-Critic architecture are still missing. 

%RL is usually based on a~set of hyperparameters that are problem-dependent and stage-of-the-process dependent. They usually can be optimized in terms of the common goal which is the maximization of future rewards. However, here we postulate that the hyperparameters that regulate the dispersion of the action distribution can not be driven by this goal. Instead, we propose to set this dispersion to be high enough to enable policy evaluation, but not higher. 

This paper combines actor-critic structure with experience replay in the old-fashioned way introduced in~\citep{2009wawrzynski}.

\section{Problem formulation} 
\label{sec:problem-formulation} 

We consider the typical RL setup \citep{2018sutton+1}. An agent operates in its environment in discrete time $t=1,2,\dots$. At time~$t$ it finds itself in a~state, $\state_t\in\stateSpace$, performs an action, $\ctrl_t\in\ctrlSpace$, receives a~reward, $r_t\in\real$, and the state changes to $\state_{t+1}$. 

In this paper, we consider the actor-critic framework \citep{1983barto+2,1998kimura+2} of RL. The goal is to optimize a~stationary control policy defined as follows. Actions are generated by a~distribution 
\Beq \label{policy} 
    \ctrl \sim \varphi(\cdot\, ; \mu, \eta) 
\Eeq
parameterized by two vectors: $\mu$ that does not affect the dispersion of the action distribution, and $\eta$ that does. Approximators produce both $\mu$ and $\eta$ 
\Beq \label{bar:mu:eta}
    \begin{split} 
    \mu & = \bar \mu (\state; \Aparam_\mu) \\
    \eta & = \bar \eta(\state; \Aparam_\eta), 
    \end{split} 
\Eeq
with $\Aparam_\mu$ and $\Aparam_\eta$ being their vectors of parameters. 

\paragraph{Neural-normal policy.}
This policy is applicable for $\ctrlSpace=\real^d$. Both $\bar\mu(\state; \Aparam_\mu)$ and $\bar\eta(\state;\Aparam_\eta)$ are neural networks with input $\state$ and weights $\Aparam_\mu$ and $\Aparam_\eta$, respectively.\footnote{They could also be implemented as a~single network outputting both $\mu$ and $\eta$, but for brevity we use this distinction.} The action is sampled from the normal distribution with mean $\bar\mu(\state; \Aparam_\mu)$ and covariance matrix $\text{diag}(\exp(2\bar\eta(\state;\Aparam_\eta)))$. Thus we denote 
\Beq 
    \sigma(\state_t;\Aparam_\eta) = \exp(\bar\eta(\state_t;\Aparam_\eta)), 
\Eeq 
and generate an action as 
\Beq
    \ctrl_t 
    = \bar\mu(\state_t;\Aparam_\mu) + \xi_t \circ \sigma(\state_t;\Aparam_\eta), 
\Eeq
where $\xi_t \sim N(0,I)$, $\sigma(\state_t;\Aparam_\eta)$ is a~vector of standard deviations for different action components and ``$\circ$'' denotes the Hadamard (elementwise) product. 

\paragraph{Experience replay.} We assume that the policy is optimized with the use of experience replay. The agent's experience, i.e., states, actions, and rewards, are stored in a~memory buffer. Simultaneously to the agent's operation in the environment, the experience is called from the buffer and used to optimize $\Aparam_\mu$ and $\Aparam_\eta$. 

\paragraph{Goal.} The vectors of weights, $\Aparam_\mu$ and $\Aparam_\eta$ define a~policy, $\pi$. The criterion of the policy optimization is the maximization of the value function 
\Beq \label{def:V} 
    V^\pi(\state) = E_\pi\left(\left.\left.\sum\right._{i\geq0} \gamma^i r_{t+i}\right| \state_t = \state\right) 
\Eeq
for each state $\state$; $\gamma\in[0,1)$ is a~coefficient -- the discount factor. 

The value function may be estimated based on so-called $n$-step returns, namely 
\Beq \label{n-step-returns} 
    V^\pi(\state_t) \cong r_t + \dots + \gamma^{n-1} r_{t+n-1} + \gamma^n V^\pi(\state_{t+n}), 
\Eeq 
for any $n\in\mathbb{N}$. 

The optimization criterion of $\Aparam_\mu$ maximizes the probability of experienced actions that led to high rewards afterward. However, $\Aparam_\eta$ should be tuned to keep a~proper balance between exploration and exploitation in the agent's behavior. 

\section{Method} 
\label{sec:method} 

\subsection{General idea} 

We consider an actor-critic algorithm with experience replay. The algorithm keeps a~window of length $M$ of previous events and continuously optimizes the policy. We postulate that the dispersion of the distribution of the current policy should be sufficient to produce actions that will enable the evaluation and selection of future policies, i.e., these actions should be likely in future policies. However, the future policies are unknown. We assume that they will be as different from the current policy as the current policy is different from those that produced the actions registered in the current memory buffer. 

\begin{figure}[h]
    \centering
    \includegraphics[width=0.7\textwidth]{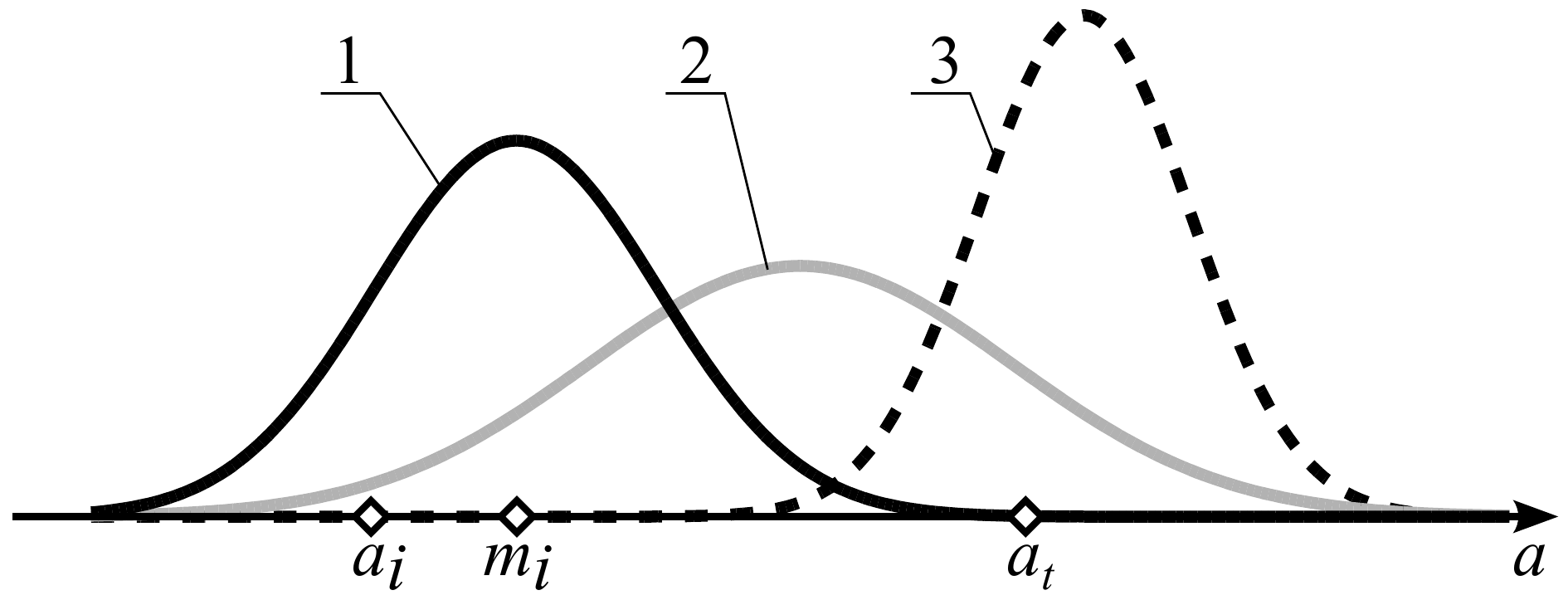} 
    \caption{Illustration. 1 -- the policy that generated action $\ctrl_i$; its mode is $m_i$, 2 -- the current policy, 3 -- a~hypothetical future policy. Under the proposed method, the dispersion of the current policy ensures that the mode and action of the previous policy are likely according to the current one. Symmetrically, the current action $\ctrl_t$ should enable the evaluation of the future policy. } 
    \label{fig:policies} 
\end{figure}

Following the above postulate, we propose two rules to optimize $\Aparam_\eta$: 
\begin{description} 
\item[R1:] 
For the $i$-th event registered in the memory buffer, the {\it mode} of the distribution that has produced the action $\ctrl_i$ should be likely according to the current policy and the state $\state_i$. 
\item[R2:]
For the same event, the {\it action} $\ctrl_i$ should be likely according to the current policy and the state $\state_i$. 
\end{description} 
The above rules (illustrated in Fig.~\ref{fig:policies}) are intended to have the following effects: 1) When the action distribution for a given state, $\state$, changes due to learning, the current action distribution for the state $\state$ is dispersed, thereby enabling to evaluate a~broad range of behaviors expected to be exercised in the course of the learning. 2) However, when the policy approaches the optimum and the pace of its change decreases, the distribution becomes less dispersed, enabling more precise action choices. Eventually, the action distribution converges to a~deterministic choice when the policy converges. 

\subsection{Operationalization} 

The mode of the distribution that has produced the action $\ctrl_i$ is defined as 
\Beq \label{def:mode} 
    m_i = \arg\max_\ctrl \varphi(\ctrl;\bar\mu(\state_i;\Aparam_\mu),\bar\eta(\state_i;\Aparam_\eta))
\Eeq
for $\Aparam_\mu$ and $\Aparam_\eta$ used at time $i$. Following the aforementioned rules R1 and R2, to optimize $\Aparam_\eta$ we minimize the loss 
\Beq \label{def:l} 
    l_i(\Aparam_\eta) =   -\ln\varphi(m_i;\bar\mu(\state_i;\Aparam_\mu),\bar\eta(\state_i;\Aparam_\eta))
    - \alpha \ln\varphi(\ctrl_i;\bar\mu(\state_i;\Aparam_\mu),\bar\eta(\state_i;\Aparam_\eta))
\Eeq
averaged over the events collected in the memory buffer. $\alpha>0$ is a coefficient. 

\paragraph{Neural-normal policy.} 
For this policy the mode \eqref{def:mode} is equal to $m_i=\bar\mu(\state_i;\Aparam_\mu)$ for $\Aparam_\mu$ applied at time $i$. The loss \eqref{def:l} then takes the form 
\Beq 
    \begin{split} 
    l_i(\Aparam_\eta) = & {\bf 1}^T \frac12 (m_i - \bar\mu(\state_i;\Aparam_\mu))^2 \circ \sigma(\state_i;\Aparam_\eta)^{-2}  
                + \alpha {\bf 1}^T \frac12 (\ctrl_i - \bar\mu(\state_i;\Aparam_\mu))^2 \circ \sigma(\state_i;\Aparam_\eta)^{-2} \\ 
                & + (1+\alpha) {\bf 1}^T \bar\eta(\state_i;\Aparam_\eta)  
                + \text{const},  
    \end{split}
\Eeq
where ${\bf1}$ is a~vector of ones (${\bf1}^Tv$ is a~sum of elements in $v$), and the squares $\cdot^2$ and $\cdot^{-2}$ are elementwise. 

\subsection{Algorithm} 

\begin{algorithm}%[b!]
\caption{Experience replay in Actor-Critic with Experience Replay and Adaptive eXploration, ACERAX}
\label{alg:ACERAX} 
\begin{algorithmic}[1]
\STATE
Sample $i \sim U(\{t-M, \dots, t-n\})$. 
\STATE
Compute the temporal difference 
%\Beq \label{gen:temp:diff}
%    \begin{split} 
\Beqso
    e = \sum_{k=0}^{n-1} & \gamma^k\left(r_{i+k} + \gamma \bar V(\state_{i+k+1};\Cparam) - \bar V(\state_{i+k};\Cparam)\right) 
    \times \\ & \times 
    \psi_b\!\left(\prod_{j=i}^{i+k} \frac{\varphi(\ctrl_j; \bar\mu(\state_j;\Aparam_\mu),\bar\eta(\state_j;\Aparam_\eta))}{\varphi_j}\right) 
\Eeqso
%    \end{split} 
%\Eeq
\STATE
Compute the Critic gradient estimate: 
$$
    \Delta\Cparam = e 
    \frac{\pd \bar V(\state_i;\Cparam)}{\pd \Cparam\T} 
$$
\STATE
Compute the Actor gradient estimate: 
$$
    \Delta\Aparam_\mu = e 
    \frac{\pd \ln \varphi(\ctrl_i;\bar\mu(\state_i;\Aparam_\mu), \bar\eta(\state_i;\Aparam_\eta))}
    {\pd \Aparam_\mu\T} - \frac{\pd p(\bar\mu(\state_i;\Aparam_\mu))}{\pd\Aparam_\mu\T}
$$
\STATE
Compute the dispersion loss gradient estimate:
$$
    \Delta\Aparam_\eta = \frac{\pd l_i(\Aparam_\eta)} {\pd\Aparam_\eta\T} 
$$
\STATE
Update $\Cparam$ with $\Delta\Cparam$, $\Aparam_\mu$ with $\Delta\Aparam_\mu$, and $\Aparam_\eta$ with $\Delta\Aparam_\eta$.
\end{algorithmic} 
\end{algorithm} 

The algorithm presented here is Actor-Critic with Experience Replay and Adaptive eXploration, ACERAX. It is based on the Actor-Critic structure shown in Section~\ref{sec:problem-formulation},
experience replay, and $n$-step returns. The algorithm uses a~critic, $\bar V(\state;\Cparam)$, an~approximator of the value function with weights $\Cparam$. 

At each time $t$ of the agent-environment interaction, the following tuple is registered
$$
    \langle \state_t, \ctrl_t, r_t, m_t, \varphi_t \rangle, \,
    \mu_t \!=\! \bar\mu(\state_t; \Aparam_\mu),  \,
    \varphi_t \!=\! \varphi(\ctrl_t; \mu_t,  \bar\eta(\state_t;\Aparam_\eta)). 
$$
As the agent-environment interaction continues, previous experience is being replayed; that is, Algorithm~\ref{alg:ACERAX} is being recurrently executed.  

In Line 1, the algorithm selects an~experienced event to replay. 
In Line~2, it determines the relative quality of~$\ctrl_i$, namely the temporal difference multiplied by a~softly truncated density ratio. $\Aparam$ is changing due to learning. Thus the conditional distribution $(\ctrl_i|\state_{i})$ is now different than it was at the time when the action $\ctrl_i$ was executed. The product of density ratios in $e$ accounts for this discrepancy in distributions. To limit the variance of the density ratios, the soft-truncating function~$\psi_b$ is applied, e.g., 
\Beq \label{psi:example}
    \psi_b(z) = b\tanh(z/b), 
\Eeq
for $b>1$. In the ACER algorithm~\citep{2009wawrzynski}, the hard truncation function, $\min\{\cdot,b\}$ is used for the same purpose, which is limiting density ratios necessary in designating updates due to action distribution discrepancies. However, soft-truncating distinguishes the magnitude of density ratio and may be expected to work slightly better than the hard truncation. 

In Line~3, an improvement direction of the parameters of critic, $\Cparam$, is computed. $\Delta\Cparam$ is designed to make $\bar V(\cdot;\Cparam)$ approximate the value function better. 

In Line~4, an improvement direction for the actor parameter $\Aparam_\mu$ is computed. The increment $\Delta\Aparam_\mu$ is designed to increase/decrease the likelihood of occurrence of the sequence of actions $\ctrl_i$ proportionally to $e$. The $p$ function is a~penalty for improper values of $\bar\mu(\state_i;\Aparam_\mu)$, e.g., exceeding a~box to which the actions should belong. 

In Line 5, an improvement direction for the actor parameter $\Aparam_\eta$ is computed, for $l_i(\Aparam_\eta)$ defined in \eqref{def:l}. 

The improvement directions $\Delta\Cparam$, $\Delta\Aparam_\mu$ and $\Delta\Aparam_\eta$ are applied in Line 6 to update $\Cparam$, $\Aparam_\mu$, and $\Aparam_\eta$, respectively, with the use of either ADAM, SGD, or another method of stochastic optimization. They may be applied in minibatches, several at a~time. 

\section{Experimental study}
\label{sec:experiments} 

This section presents simulations whose purpose is to evaluate the ACERAX algorithm introduced in Sec.~\ref{sec:method}. In our first experiment, we determine the algorithm's sensitivity to its $\alpha$ parameter and its approximately optimal value across several RL problems. In the second experiment, we look at action standard deviations the algorithm determines and compare them with constant action standard deviations optimized manually. We call the algorithm with constant approximately optimal action standard deviations ACER, as it differs only in details from that presented \citep{2009wawrzynski} under that name. In the third experiment, we compare ACERAX to two state-of-the-art RL algorithms: the Soft Actor-Critic (SAC) \citep{2018haarnoja+3} algorithm and the Proximal Policy Optimization (PPO) \citep{2017schulman+4} algorithm. We use the Stable Baselines3 implementation~\citep{2020raffin+1} of SAC and PPO in the simulations. Our experimental software is available online.\footnote{{\it provided in the final version of the paper}}

\begin{figure*}[h!]
    \begin{tabular}{c c c c}
    \includegraphics[width=0.23\textwidth]{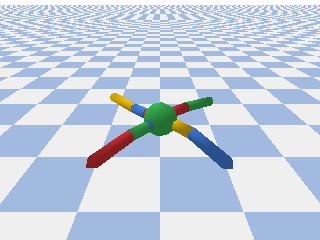} &
    \includegraphics[width=0.23\textwidth]{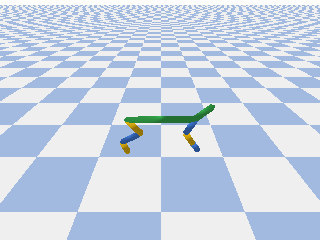} &
    \includegraphics[width=0.23\textwidth]{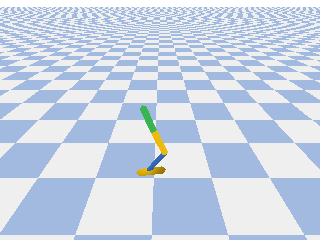} &
    \includegraphics[width=0.23\textwidth]{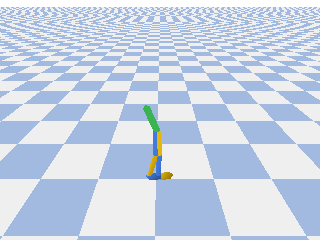}
    \end{tabular}
    \caption{Environments  used  in  simulations. From the left:  Ant,  HalfCheetah, Hopper, Walker2D.}
    \label{fig:Envs}
\end{figure*}

For the comparison of the RL algorithms to be the most informative, we chose four challenging tasks inspired by robotics, namely Ant, Hopper, HalfCheetah, and Walker2D (see Fig.~\ref{fig:Envs}) from the PyBullet physics simulator \citep{2019coumans}.
%A~simulator that is more popular in the RL community is MuJoCo \citep{2012todorov}.\footnote{We chose PyBullet because it is freeware under the zlib license, while MuJoCo is commercial software.} 

%applied to the environments considered in MuJoCo are well known. However, PyBullet environments introduce several changes to MuJoCo tasks, which make them more realistic and thus more difficult. Additionally, physics in MuJoCo and PyBullets differ slightly \citep{2015erez+2}, hence we needed to tune the hyperparameters. 
%For each learning algorithm, we use actor and critic structures as described in \citep{2018haarnoja+3}. That is, both structures have the form of neural networks with two hidden layers of 256 units each. \pawel{TODO}

%We train agents using desktop PCs with Intel\texttrademark Core i5-8400 CPU (6×3.8GHz) and 32GB RAM.

\subsection{Experimental settings}

Each learning run lasted for 3 million timesteps. Every 30000 timesteps, the training was paused, and a~simulation was made with frozen weights and without exploration for 5 test episodes. An average sum of rewards within a~test episode was registered. Each run was repeated five times. 

We have taken implementation SAC and PPO from \citep{2021raffin+5}, and the hyperparameters that assure the state-of-the-art performance of these algorithms from \citep{2020raffin+1}. For each environment, hyperparameters for ACER and ACERAX, such as step-sizes, were optimized to yield the highest ultimate average rewards. The values of these hyperparameters are reported in the appendix.

\subsection{Searching for $\alpha$}

\begin{figure*}[h!]
    \begin{tabular}{c c}
    \includegraphics[width=0.5\textwidth]{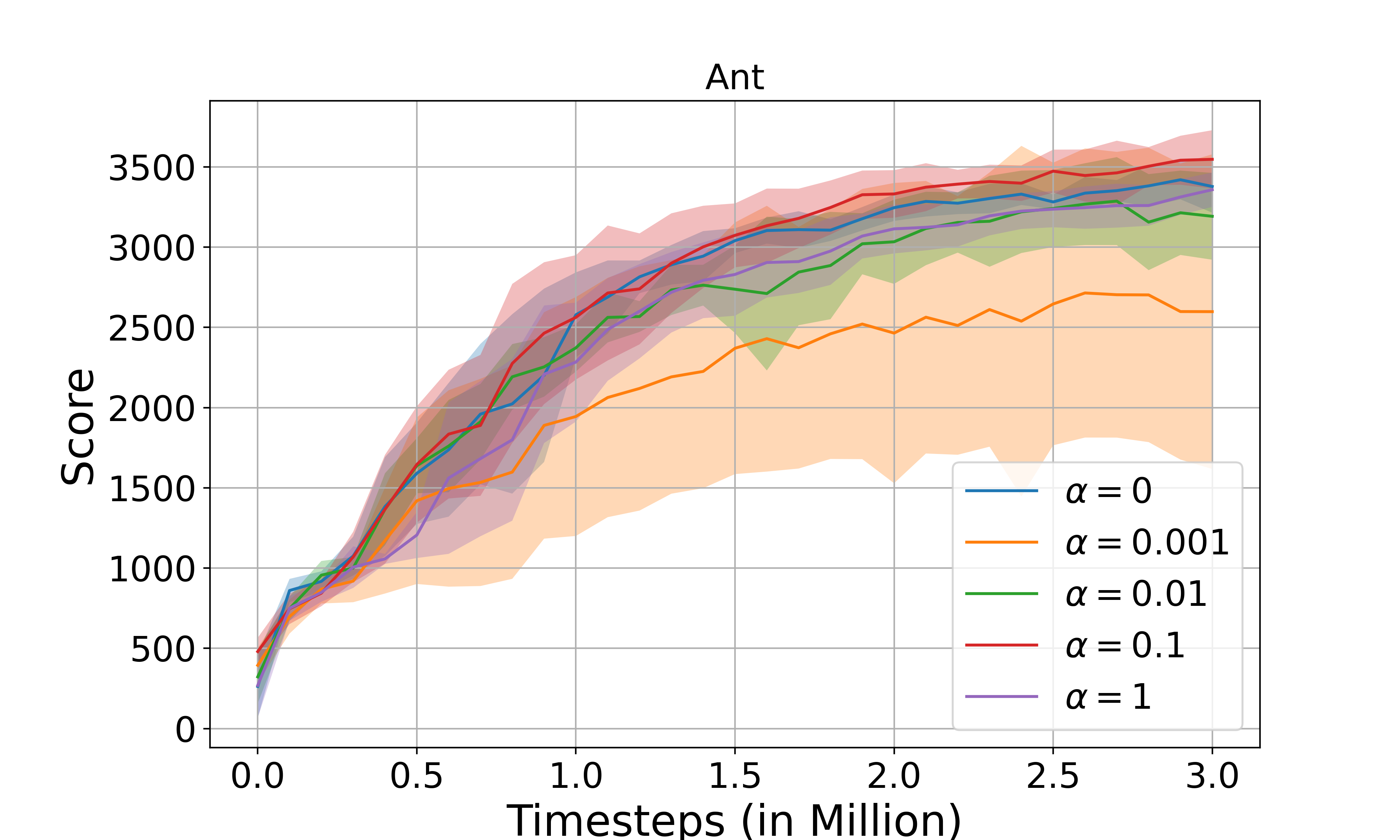} &\!\!\!\!\!
    \includegraphics[width=0.5\textwidth]{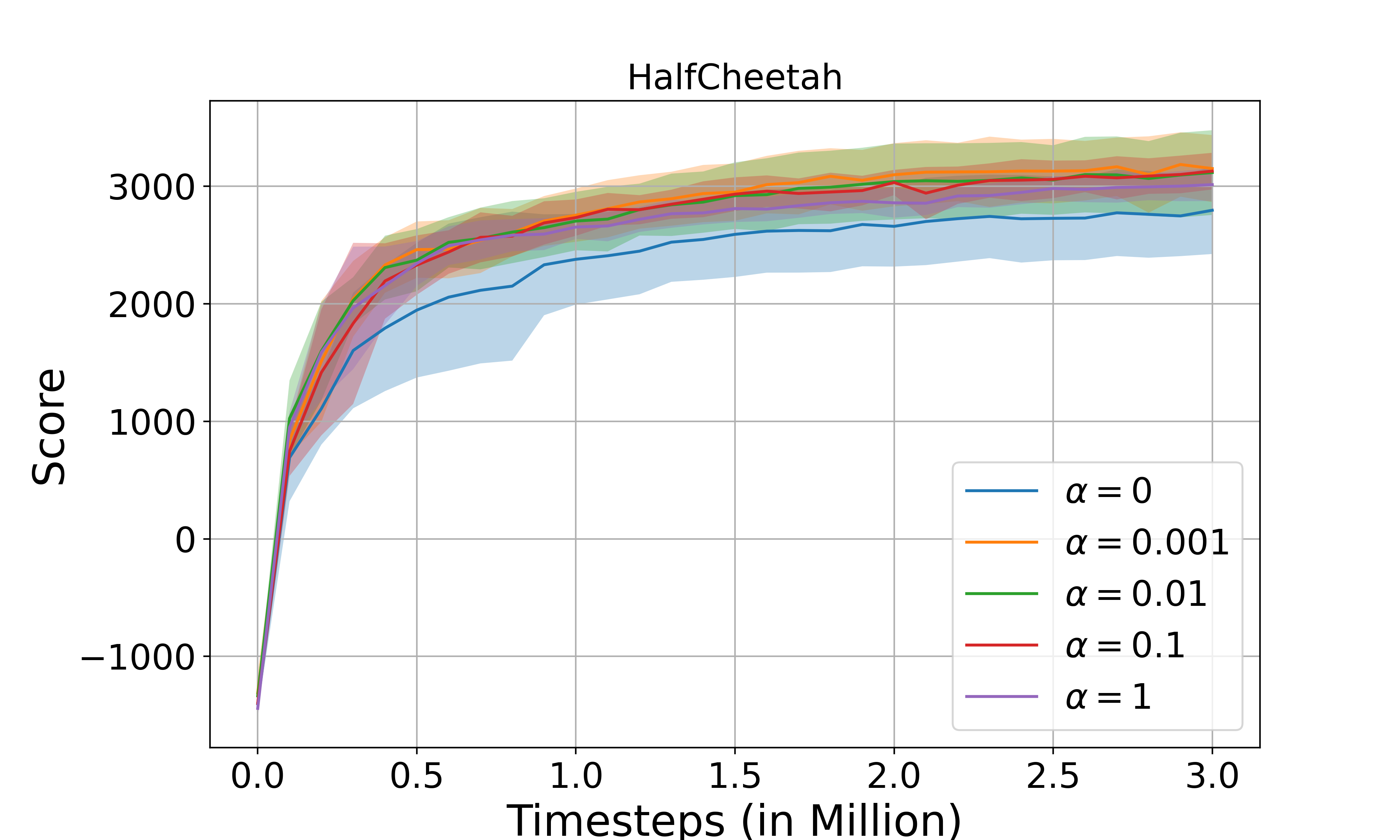} \\%&\!\!\!\!\!
    \includegraphics[width=0.5\textwidth]{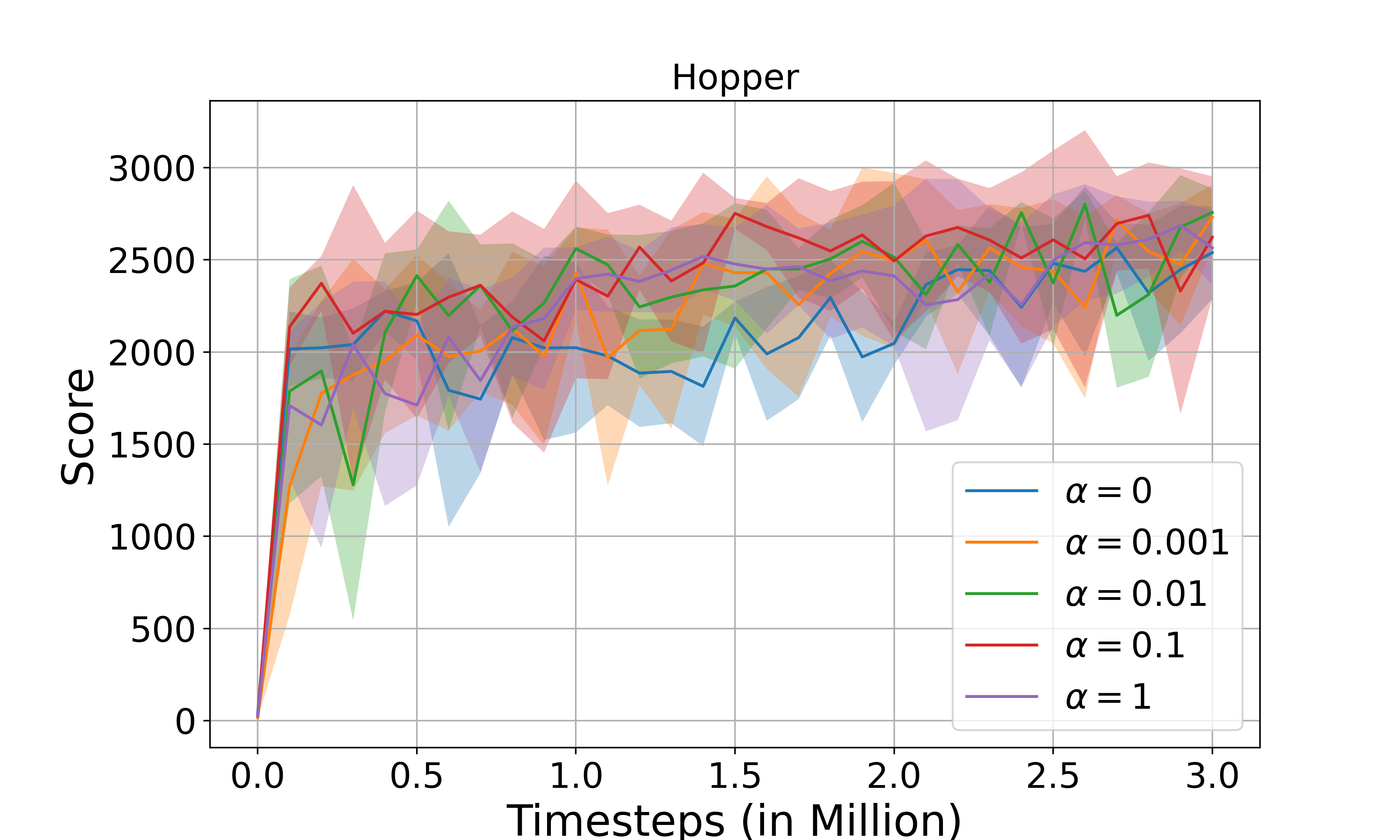} &\!\!\!\!\!
    \includegraphics[width=0.5\textwidth]{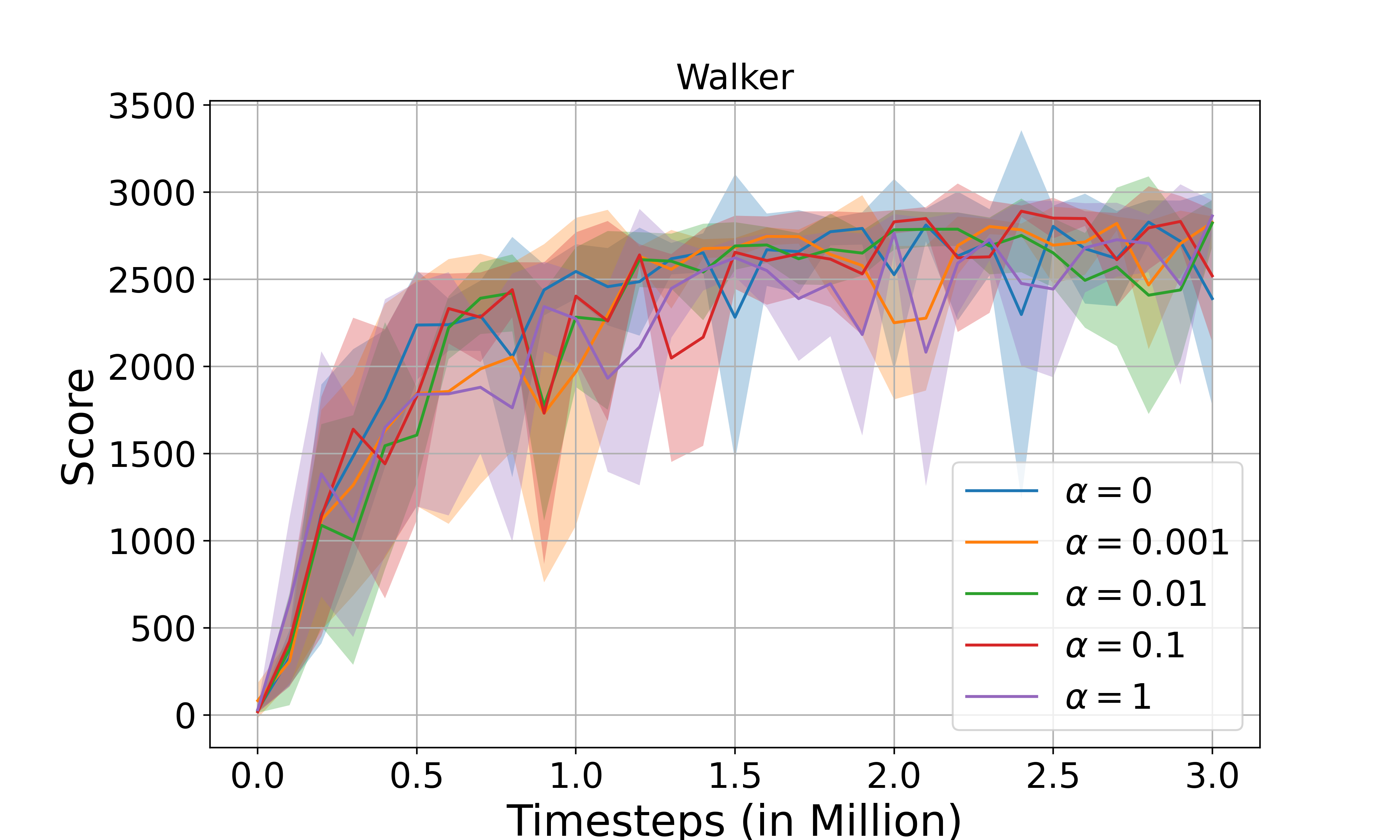}
    \end{tabular}
    \caption{Results of ACERAX and different $\alpha$. From the left:  Ant, HalfCheetah, Hopper, Walker2D.}
    \label{fig:ACERAX:alpha}
\end{figure*}

For each environment, we tried $\alpha=0, 0.001, 0.01, 0.1, 1$. The results are depicted in Fig.~\ref{fig:ACERAX:alpha}. It is seen that the learning is not very sensitive to this parameter. $\alpha=0$ means that the previous modes of action distributions and the previous actions are made likely according to the current policy. $\alpha>0$ means that the previous actions are made likely, which is intended to be a~form of regularization: It prevents the policy from degrading to a~deterministic one too early. The experiments suggest that performance is almost insensitive to this parameter when $\alpha\in[0,1]$. We choose $\alpha=0.1$ for the rest of the experiments. 

%We choose $\alpha=0.001$ for HalfCheetah environment and $\alpha=0.1$ for the others. 

\subsection{Different initial action standard deviations} 

\begin{figure*}[h!]
    \begin{tabular}{c c}
    \includegraphics[width=0.5\textwidth]{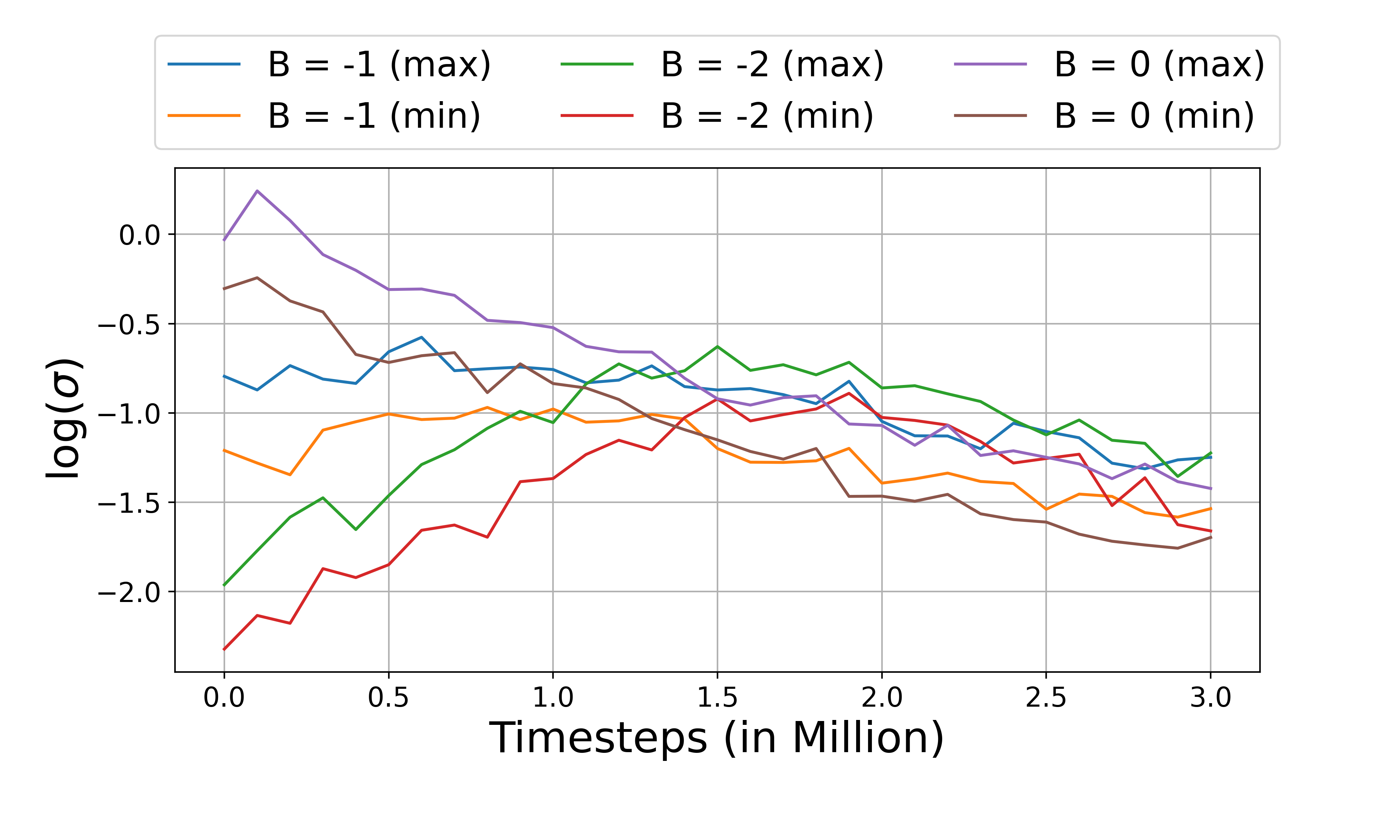} &\!\!\!\!\!
    \includegraphics[width=0.5\textwidth]{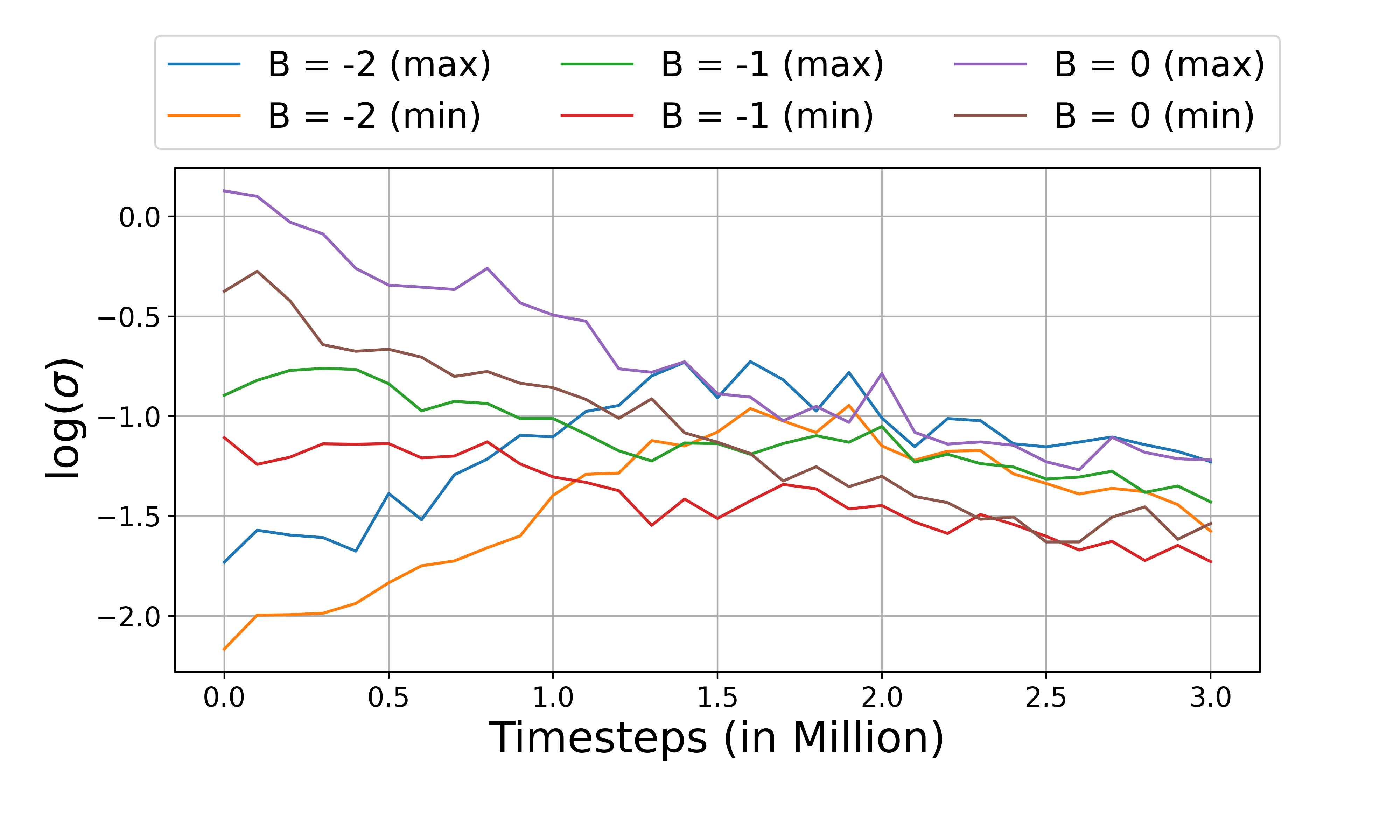} \\%&\!\!\!\!\!
    \includegraphics[width=0.5\textwidth]{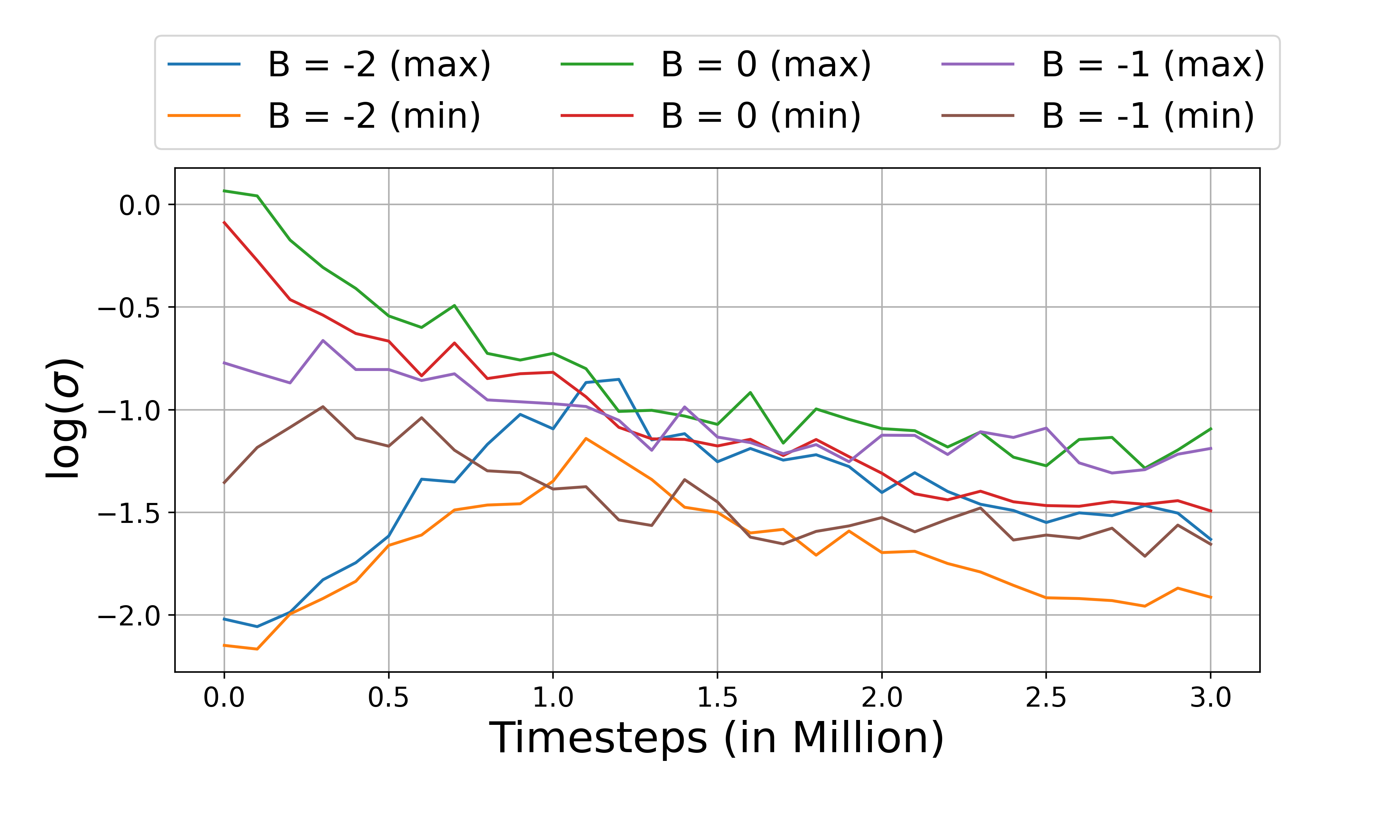} &\!\!\!\!\!
    \includegraphics[width=0.5\textwidth]{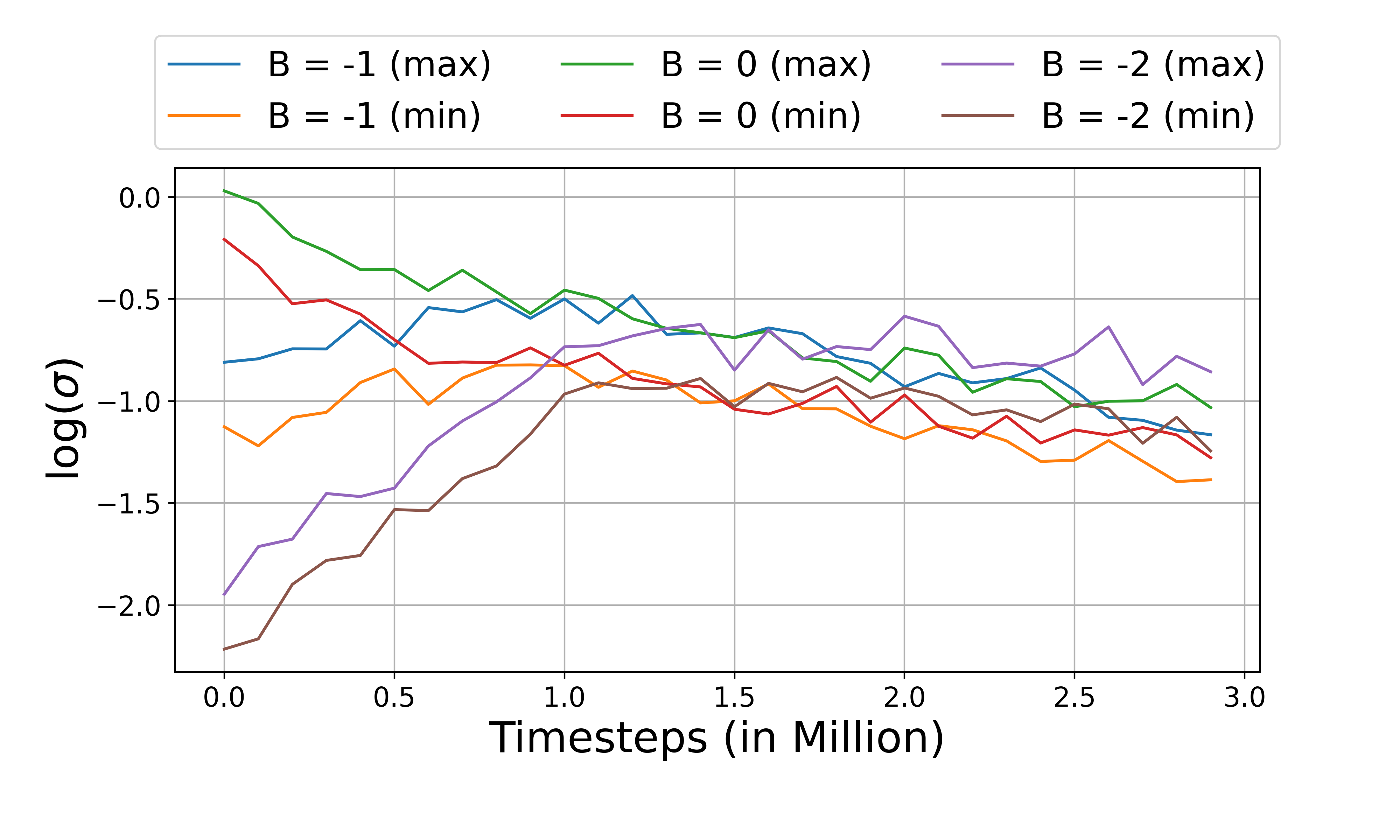}
    \end{tabular}
    \caption{Comparison of the behavior of $\eta = \log (\sigma)$ during training for different initial biases in the last layer of the $\bar\eta$ network.
Each plot represents the maximum or minimum value of the mean calculated over the coordinates of the standard deviation vector. The averages were calculated based on five independent runs. From the left:  Ant, HalfCheetah, Hopper, Walker2D.}
    \label{fig:Eta0}
\end{figure*}

We have optimized in a~grid search constant standard deviations for actions for ACER in these environments. In all environments we selected $\sigma=0.4\approx\exp(-1.2)$. In another experiment, we verify the average outputs of the $\bar\eta$ network depending on its initialization. To this end, we impose different initial biases in the output neurons of this network. Afterward, we register the outputs of this network over time. Fig.~\ref{fig:Eta0} presents the trajectories of average $\min_i\bar\eta_i(\state_t;\Aparam_\eta)$ and $\max_i\bar\eta_i(\state_t;\Aparam_\eta)$. It is seen that regardless of the initialization, these outputs converge quite close to $-1.2$. We also see that $\min_i\bar\eta_i(\state_t;\Aparam_\eta)<\max_i\bar\eta_i(\state_t;\Aparam_\eta)$ which justifies the need of designating standard deviations of different action dimensions separately. 
%\wojtek{It sounds like a major limitation of other methods, \textit{i.e.}, they use the same value of dispersion for all the action dimensions. Shouldn't we highlight this advantage of our method?} PW: unfortunatelly, other methods also distinguish different dimensions of action 

\subsection{Comparison of ACERAX, ACER, PPO, and SAC}

\begin{figure*}[h!]
    \begin{tabular}{c c}
    \includegraphics[width=0.5\linewidth]{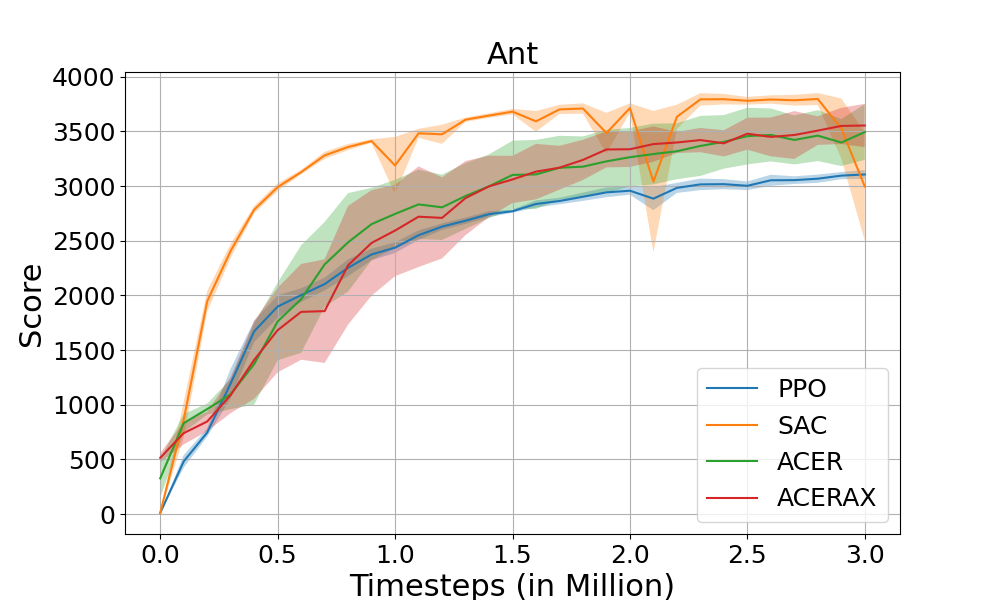} &\!\!\!\!\!
    \includegraphics[width=0.5\linewidth]{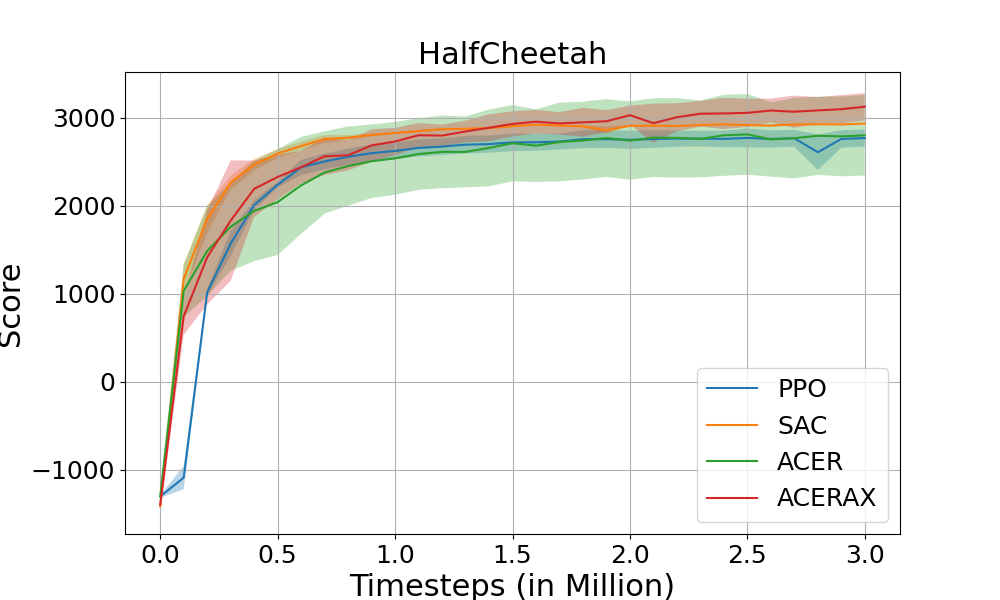} \\%&\!\!\!\!\!
    \includegraphics[width=0.5\linewidth]{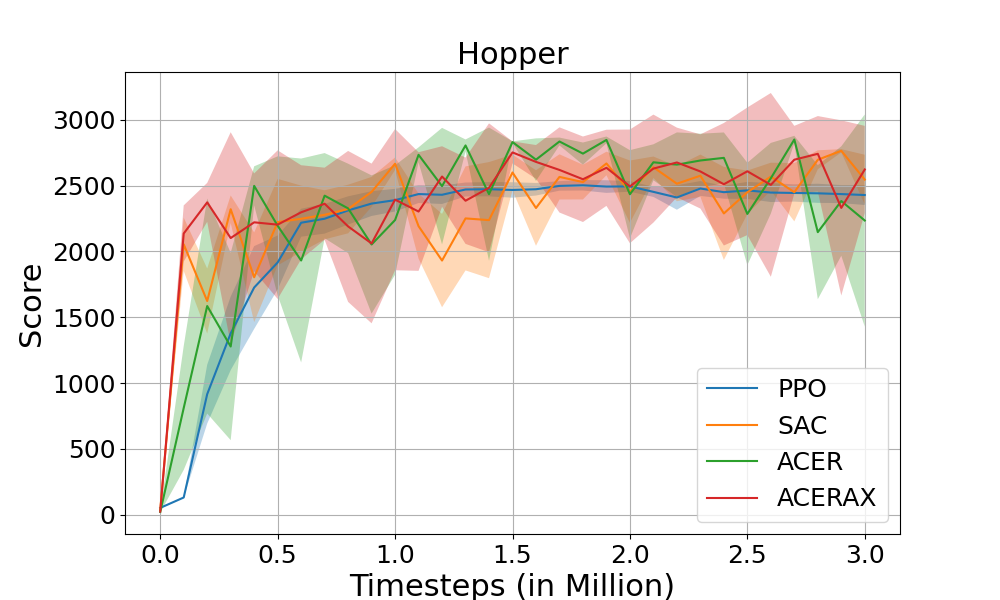} &\!\!\!\!\! 
    \includegraphics[width=0.5\linewidth]{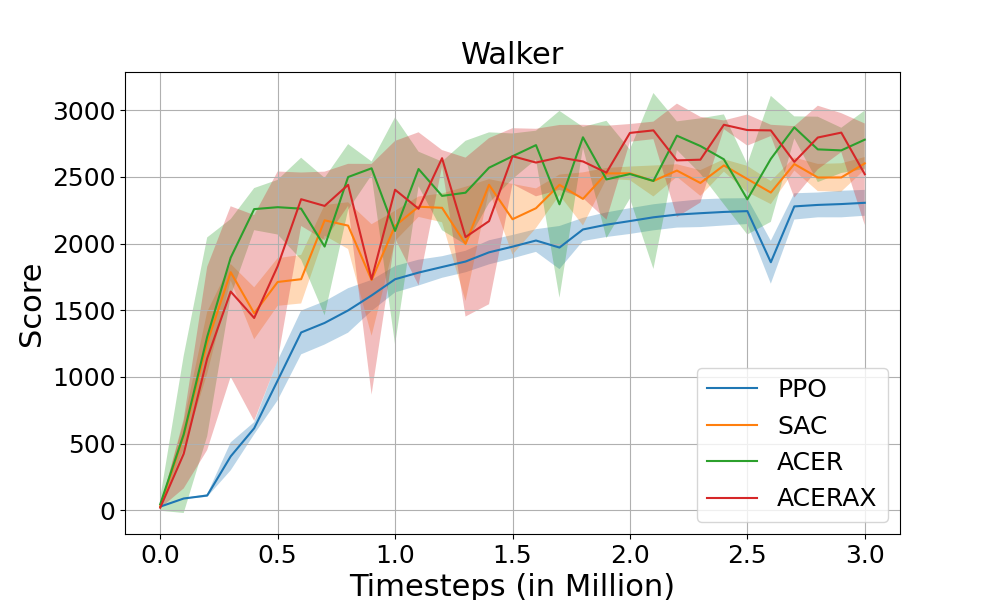}
    \end{tabular}
    \caption{Results for ACERAX, ACER, PPO, and SAC. From the left:  Ant, HalfCheetah, Hopper, Walker2D.}
    \label{fig:AllAlgos}
\end{figure*}

Fig. \ref{fig:AllAlgos} presents learning curves for all four environments for SAC, PPO, ACER, and our proposed ACERAX. It is seen that all algorithms exhibit a similar speed of learning and ultimate rewards, with ACERAX performing very well according to both these criteria. (In fact, ACERAX yields the best average ultimate performance in 3 out of 4 environments, but this result is not statistically significant.) However, the reference algorithms control exploration-exploitation balance with heuristic or problem-dependent coefficients. In ACERAX, this balance is based on an an~universal principle that is somehow affected by the $\alpha$ coefficient but is not very sensitive to this coefficient. 

\subsection{Discussion} 
\label{sec:discussion}

Exploration is necessary for reinforcement learning, but it inevitably deteriorates currently expected rewards. Best previous approaches to tuning the amount of exploration online are based on additional rewards for action distribution entropy. But the scale of these rewards is generally problem-dependent, as this is never known in advance what fraction of the original rewards need be traded for action distribution entropy to assure proper exploration. 

The approach analyzed here is based on the assumption that the policy is optimized on the data in the replay buffer. The current action distribution dispersion should be sufficient to support policy evaluation and selection when this action will be replayed from the buffer in the future. 

Our approach generally yielded good performance in the experiments. That happened at the cost of an additional neural network, $\bar\eta$, that controlled the action distribution dispersion. Our experiments suggest that this network should be much smaller than the $\bar\mu$ network -- it had ten times fewer neurons in each layer. We interpret this latter condition as follows: It should be impossible for the $\bar\eta$ network to overfit to the experience, as it would result in nearly zero action distribution dispersion for some states. 

%It is unknown what the policy is going to be in the future. The assumption that we make is that the pace of the policy change will be preserved. Therefore, the dispersion of the distribution of the current actions makes them as likely in the future as previous actions are likely now. This assumption allows us to define a~criterion for the action distribution dispersion that is independent of the rewards. Our experiments positively verified the above assumption. The actions performed do enable evaluation of any current policy and their distribution dispersion generally decreased as the policy converged. That in turn led to the very good ultimate performance.  

\section{Conclusions} 
\label{sec:conclusions} 

Exploration/exploitation balance is a~fundamental problem in reinforcement learning. In this paper, we analyzed an approach to the adaptive designation of the amount of randomness in action distribution. In this approach, the probability densities are maximized by the modes of the distribution of actions in the replay buffer. Consequently, current actions are likely to support the evaluation and selection of future policies. Furthermore, this strategy diminishes the randomness in actions when the policy converges, giving the agent increasing control over its actions. The RL algorithm based on this strategy introduced here, ACERAX, was verified on four challenging robotic-like benchmarks: Ant, HalfCheetah, Hopper, and Walker2D, with good results. Our method makes the action standard deviations converge to values similar to those resulting from trial-and-error optimization. 

Our proposed strategy for optimizing dispersion of action distribution %is versatile and in principle can be adopted to all reinforcement learning algorithms based on the actor-critic framework, including PPO and SAC. 
%The strategy proposed here 
is based on the maximization of weighted logarithms of previous modes of action distributions and previous actions. The optimal weights are potentially problem-dependent, although the strategy appears to be barely sensitive to these weights. Getting rid of any potentially problem-dependent coefficients from this strategy is a~curious direction for further research. 

%\wojtek{The last sentence of the abstract is: " The above principle is verified here on challenging benchmarks, namely Ant, HalfCheetah, Hopper, and Walker2D, with good results." It's a detail, but I would unify our opinion of the quality of obtained results throughout the paper :D}

\wyciete{
\section*{Acknowledgments}

This work was partially funded by a grant from the Warsaw University of Technology Scientific Discipline Council for Computer Science and Telecommunications.
}

\ifdefined\headicml
\bibliographystyle{icml2022}
\fi 
\ifdefined\headijcai
\bibliographystyle{named}
\fi
\ifdefined\headneurips
\fi

%\bibliography{references}

\begin{thebibliography}{}

\bibitem[Barto et~al., 1983]{1983barto+2}
Barto, A.~G., Sutton, R.~S., and Anderson, C.~W. (1983).
\newblock Neuronlike adaptive elements that can learn difficult learning
  control problems.
\newblock {\em IEEE Transactions on Systems, Man, and Cybernetics B},
  13:834--846.

\bibitem[Coumans and Bai, 2019]{2019coumans}
Coumans, E. and Bai, Y. (2016--2019).
\newblock Pybullet, a python module for physics simulation for games, robotics
  and machine learning.
\newblock \url{http://pybullet.org}.

\bibitem[Haarnoja et~al., 2018]{2018haarnoja+3}
Haarnoja, T., Zhou, A., Abbeel, P., and Levine, S. (2018).
\newblock Soft actor-critic: Offpolicy maximum entropy deep reinforcement
  learning with a stochastic actor.
\newblock arXiv:1801.01290.

\bibitem[Haarnoja et~al., 2019]{2019haarnoja+many}
Haarnoja, T., Zhou, A., Hartikainen, K., Tucker, G., Ha, S., Tan, J., Kumar,
  V., Zhu, H., Gupta, A., Abbeel, P., and Levine, S. (2019).
\newblock Soft actor-critic algorithms and applications.
\newblock arXiv:1812.05905.

\bibitem[Hong et~al., 2018]{2018hong+many}
Hong, Z.-W., Shann, T.-Y., Su, S.-Y., Chang, Y.-H., Fu, T.-J., and Lee, C.-Y.
  (2018).
\newblock Diversity-driven exploration strategy for deep reinforcement
  learning.
\newblock In {\em Neural Information Processing Systems (NeurIPS)}.

\bibitem[Jaynes, 1957]{1957jaynes}
Jaynes, E.~T. (1957).
\newblock Information theory and statistical mechanics. ii.
\newblock {\em Physical Review}, 108:171--190.

\bibitem[Kakade and Langford, 2002]{2002kakade+1}
Kakade, S. and Langford, J. (2002).
\newblock Approximately optimal approximate reinforcement learning.
\newblock In {\em International Conference on Machine Learning (ICML)}, pages
  267--274.

\bibitem[Kimura and Kobayashi, 1998]{1998kimura+2}
Kimura, H. and Kobayashi, S. (1998).
\newblock An analysis of actor/critic algorithms using eligibility traces:
  Reinforcement learning with imperfect value function.
\newblock In {\em International Conference on Machine Learning (ICML)}.

\bibitem[Lillicrap et~al., 2016]{2016lillicrap+7}
Lillicrap, T.~P., Hunt, J.~J., Pritzel, A., Heess, N., Erez, T., Tassa, Y.,
  Silver, D., and Wierstra, D. (2016).
\newblock Continuous control with deep reinforcement learning.
\newblock In {\em ICML}.

\bibitem[Mahadevan and Connell, 1992]{1992mahadevan+1}
Mahadevan, S. and Connell, J. (1992).
\newblock Automatic programming of behavior based robots using reinforcement
  learning.
\newblock {\em Artificial Intelligence}, 55(2--3):311--365.

\bibitem[Mnih et~al., 2016]{2016mnih+many}
Mnih, V., Badia, A.~P., Mirza, M., Graves, A., Lillicrap, T.~P., Harley, T.,
  Silver, D., and Kavukcuoglu, K. (2016).
\newblock Asynchronous methods for deep reinforcement learning.
\newblock arXiv:1602.01783.

\bibitem[Mnih et~al., 2013]{2013mnih+6}
Mnih, V., Kavukcuoglu, K., Silver, D., Graves, A., Antonoglou, I., Wierstra,
  D., and Riedmiller, M. (2013).
\newblock Playing atari with deep reinforcement learning.
\newblock arXiv:1312.5602.

\bibitem[Pathak et~al., 2017]{2017pathak+3}
Pathak, D., Agrawal, P., Efros, A.~A., and Darrell, T. (2017).
\newblock Curiosity-driven exploration by self-supervised prediction.
\newblock arXiv:1705.05363.

\bibitem[Raffin et~al., 2021]{2021raffin+5}
Raffin, A., Hill, A., Gleave, A., Kanervisto, A., Ernestus, M., and Dormann, N.
  (2021).
\newblock Stable-baselines3: Reliable reinforcement learning implementations.
\newblock {\em Journal of Machine Learning Research}, 22(268):1--8.

\bibitem[Raffin and Stulp, 2020]{2020raffin+1}
Raffin, A. and Stulp, F. (2020).
\newblock Generalized state-dependent exploration for deep reinforcement
  learning in robotics.
\newblock arXiv:2005.05719.

\bibitem[Schulman et~al., 2015]{2015schulman+4}
Schulman, J., Levine, S., Moritz, P., Jordan, M.~I., and Abbeel, P. (2015).
\newblock Trust region policy optimization.
\newblock arXiv:1502.05477.

\bibitem[Schulman et~al., 2017]{2017schulman+4}
Schulman, J., Wolski, F., Dhariwal, P., Radford, A., and Klimov, O. (2017).
\newblock Proximal policy optimization algorithms.
\newblock arXiv:1707.06347.

\bibitem[Stadie et~al., 2015]{2015stadie+3}
Stadie, B., Levine, S., and Abbeel, P. (2015).
\newblock Incentivizing exploration in reinforcement learning with deep
  predictive models.
\newblock arXiv:1507.00814.

\bibitem[Sutton and Barto, 2018]{2018sutton+1}
Sutton, R.~S. and Barto, A.~G. (2018).
\newblock {\em Reinforcement Learning: An Introduction. Second edition}.
\newblock The MIT Press.

\bibitem[Tang et~al., 2017]{2017tang+many}
Tang, H., Houthooft, R., Foote, D., Stooke, A., Chen, X., Duan, Y., Schulman,
  J., {De Turck}, F., and Abbeel, P. (2017).
\newblock \#{E}xploration: A study of count-based exploration for deep
  reinforcement learning.
\newblock arXiv:1611.04717.

\bibitem[Wang and Ni, 2020]{2020wang+1}
Wang, Y. and Ni, T. (2020).
\newblock Meta-sac: Auto-tune the entropy temperature of soft actor-critic via
  metagradient.
\newblock arXiv:2007.01932.

\bibitem[Wang et~al., 2016]{2016wang+6}
Wang, Z., Bapst, V., Heess, N., Mnih, V., Munos, R., Kavukcuoglu, K., and {de
  Freitas}, N. (2016).
\newblock Sample efficient actor-critic with experience replay.
\newblock arXiv:1611.01224.

\bibitem[Wawrzyński, 2009]{2009wawrzynski}
Wawrzyński, P. (2009).
\newblock Real-time reinforcement learning by sequential actor–critics and
  experience replay.
\newblock {\em Neural Networks}, 22(10):1484--1497.

\bibitem[Williams and Peng, 1991]{1991williams+1}
Williams, R. and Peng, J. (1991).
\newblock Function optimization using connectionist reinforcement learning
  algorithms.
\newblock {\em Connection Science}, 3(3):241--268.

\bibitem[Ziebart et~al., 2008]{2008ziebart+1}
Ziebart, B.~D., Maas, A., Bagnell, J.~A., and Dey, A.~K. (2008).
\newblock Maximum entropy inverse reinforcement learning.
\newblock In {\em AAAI Conference on Artificial Intelligence}, pages
  1433--1438.

\end{thebibliography}

%\input{neurips_checklist}

\appendix
\section{Algorithms' hyperparameters} 
\label{AlgorithmsHyperparams}

Hyperparameters of SAC and PPO have been taken from \cite{2020raffin+1}. They are presented in Tab.~\ref{tab:params:SAC} and \ref{tab:params:PPO}, respectively. 

In all experiments we used a~discount factor of $0.98$. Common hyperpameters of ACER and ACERAX are presented in Tab.~\ref{tab:params:ACER}. When possible, we have adopted hyperparameters of these algorithms from SAC. These include the actor and critic sizes and the parameters that define the intensity of experience replay. Step-sizes have been selected from the set
$$
\{\dots, 10^{-5}, 3\cdot10^{-5}, 10^{-4}, 3\cdot10^{-4}, \dots\}. 
$$
Standard deviations for actions components in ACER have been selected from the set 
$$
    \{0.1, 0.2, 0.3, 0.4, 0.5\}. 
$$
Different action components for the same environment have the same standard deviation. For all environment its selected value was the same, $0.4$. 

The $\bar\eta$ network used in the ACERAX algorithm had 40 and 30 neurons in its two hidden layers, i.e., it was of size $\langle40,30\rangle$. The step-sizes for this network are presented in Table~\ref{tab:params:ACERAX}. Biases of the $\bar\eta$ network output layer were initialized with $-1$. That means that the initial standard deviation of action components was  $\exp(-1)\approx0.37$. 

\begin{minipage}[c]{0.5\textwidth}
    \centering
    \begin{tabular}{c|c}
        \hline
        Parameter & Value \\
        \hline
        Actor size & $\langle400,300\rangle$ \\ 
        Critic size & $\langle400,300\rangle$ \\ 
        Discount factor $\gamma$ & 0.98 \\
        Replay buffer size & $3\cdot10^5$ \\
        Minibatch size & 256 \\
        %Target smoothing coef. $\tau$ & 0.005 \\
        Entropy coefficient & auto \\ 
        Target entropy & $-\dim(\mathcal{A})$ \\ 
        Learning start & $10^4$ \\
        Gradient steps & 8 \\
        Train frequency & 8 \\ 
        Step-size & $7.3\cdot10^{-4}$ \\
        Initial $\log\sigma$ & $-3$ \\
        \hline
    \end{tabular}
    \captionof{table}{SAC hyperparameters. The actor and critic sizes define numbers of neurons in hidden layers of respective neural networks. }
    \label{tab:params:SAC}
    \vspace{2em} 
    \begin{tabular}{c|c}
        \hline
        Parameter & Value \\
        \hline
        Actor size & $\langle256,256\rangle$ \\ 
        Critic size & $\langle256,256\rangle$ \\ 
        Discount factor $\gamma$ & 0.99 \\
        GAE parameter $\lambda$ & 0.9 \\
        Minibatch size & 128 \\
        Horizon & 512 \\
        Number of epochs & 20 \\
        %Value function clipping coef. & 10 \\
        %Target KL & 0.01 \\
        Step-size & $3\cdot10^{-5}$ \\
        Entropy param. & 0.0 \\ 
        Clip range & 0.4 \\
        %\multicolumn{1}{l|}{Ant and Hopper} & \\ 
        %initial log $\sigma$ & -2 \\ 
        %\multicolumn{1}{l|}{HalfCheetah and Walker2D} & \\ 
        Initial log $\sigma$ & $-2$ \\ 
        \hline
    \end{tabular}
    \captionof{table}{PPO hyperparameters.}
    \label{tab:params:PPO}
\end{minipage}
\begin{minipage}[c]{0.5\textwidth}
    \centering
    \begin{tabular}{c|c}
        \hline
        Parameter & Value \\
        \hline
        Actor size & $\langle400,300\rangle$ \\ 
        Critic size & 
        $\langle400,300\rangle$ \\ 
        $n$ & 10 \\ 
        $b$ & 3 \\
        Memory size & $10^6$\\
        Minibatch size & 256 \\
        Gradient steps & 1 \\
        Discount factor $\gamma$ & 0.98 \\
        \multicolumn{1}{l|}{Ant} \\ 
        Actor step-size & $10^{-5}$ \\ 
        Critic step-size & $10^{-5}$ \\ 
        \multicolumn{1}{l|}{HalfCheetah} \\ 
        Actor step-size & $3\cdot10^{-5}$ \\ 
        Critic step-size & $3\cdot10^{-4}$ \\ 
        \multicolumn{1}{l|}{Hopper} \\ 
        Actor step-size & $3\cdot10^{-5}$ \\ 
        Critic step-size & $3\cdot10^{-4}$ \\ 
        \multicolumn{1}{l|}{Walker2D} \\ 
        Actor step-size & $3\cdot10^{-5}$ \\ 
        Critic step-size & $3\cdot10^{-4}$ \\ 
        \hline
    \end{tabular}
    \captionof{table}{ACER and ACERAX hyperparameters.}
    \label{tab:params:ACER}
    \vspace{2em}
    \begin{tabular}{c|c}
        \hline
        Parameter & Value \\
        \hline
        $\bar\eta$ size & $\langle40,30\rangle$ \\ 
        Initial $\bar\eta$ output bias & -1 \\
        Ant $\bar\eta$ step-size & $3\cdot10^{-8}$ \\ 
        HalfChetah $\bar\eta$ step-size & $3\cdot10^{-8}$ \\ 
        Hopper $\bar\eta$ step-size & $3\cdot10^{-8}$ \\ 
        Walker2D $\bar\eta$ step-size & $3\cdot10^{-8}$ \\ 
        \hline
    \end{tabular}
    \captionof{table}{ACERAX hyperparameters.}
    \label{tab:params:ACERAX}
\end{minipage}

\comment{
\begin{table}
    \centering
    \begin{tabular}{c|c}
        \hline
        Parameter & Value \\
        \hline
        \multicolumn{1}{l|}{Ant} \\ 
        Action std. dev. & $0.4$ \\ 
        \multicolumn{1}{l|}{HalfCheetah} \\ 
        Action std. dev. & $0.4$ \\ 
        \multicolumn{1}{l|}{Hopper} \\ 
        Action std. dev. & $0.2$ \\ 
        \multicolumn{1}{l|}{Walker2D} \\ 
        Action std. dev. & $0.4$ \\ 
        \hline
    \end{tabular}
    \caption{ACER: Standard deviations for actions.}
    \label{tab:params:ACER:std}
\end{table}
}

\section{Computational resources} 

At the stage of code debugging, we used an external cluster. For the actual experimental study we used a PC equipped with AMD\texttrademark Ryzen\texttrademark Threadripper\texttrademark 1920X for about 7 weeks.

\end{document}